\documentclass[letterpaper]{article} 
\usepackage{aaai25}  
\usepackage{times}  
\usepackage{helvet}  
\usepackage{courier}  
\usepackage[hyphens]{url}  
\usepackage{graphicx} 
\usepackage{natbib}  
\usepackage{caption} 
\usepackage{algorithm}
\usepackage{algorithmic}
\usepackage{newfloat}
\usepackage{listings}
\usepackage[accsupp]{axessibility}
\usepackage{makecell}
\usepackage{multirow}
\usepackage{amsmath}
\usepackage{amssymb}
\usepackage{booktabs}
\usepackage{subcaption}
\usepackage{cleveref}
\usepackage{xspace}
\usepackage{paralist}
\usepackage{xcolor}

\DeclareCaptionStyle{ruled}{labelfont=normalfont,labelsep=colon,strut=off} 
\floatstyle{ruled}
\newfloat{listing}{tb}{lst}{}
\floatname{listing}{Listing}
\title{All-in-One: Transferring Vision Foundation Models into Stereo Matching}
\author{
    Jingyi Zhou\textsuperscript{\rm 1}\equalcontrib,
    Haoyu Zhang\textsuperscript{\rm 1}\equalcontrib,
    Jiakang Yuan\textsuperscript{\rm 1}\equalcontrib,
    Peng Ye\textsuperscript{\rm 1,3,4}$^\dagger$,
    Tao Chen\textsuperscript{\rm 1}\thanks{Corresponding authors.},
    Hao Jiang\textsuperscript{\rm 2},
    Meiya Chen\textsuperscript{\rm 2},
    Yangyang Zhang\textsuperscript{\rm 2}
}

\affiliations{
    \textsuperscript{\rm 1}School of Information Science and Technology, Fudan University, China\\

    \textsuperscript{\rm 2}Xiaomi Inc., Beijing, China\\
    \textsuperscript{\rm 3}Shanghai AI Laboratory, Shanghai, China\\
    \textsuperscript{\rm 4}The Chinese University of Hong Kong
}

\usepackage{bibentry}

\begin{document}

\maketitle

\begin{abstract}
    As a fundamental vision task, stereo matching 
  has made remarkable progress. While recent iterative optimization-based methods have achieved promising performance, their feature extraction capabilities still have room for improvement. 
  Inspired by the ability of vision foundation models (VFMs) to extract general representations, in this work, we propose AIO-Stereo which can flexibly select and transfer knowledge from multiple heterogeneous VFMs 
  to a single stereo matching model. To better reconcile features between heterogeneous VFMs and the stereo matching model and fully exploit prior knowledge from VFMs, we proposed a dual-level feature utilization mechanism that aligns heterogeneous features
  and transfers multi-level knowledge. Based on the mechanism, a dual-level selective knowledge transfer module is designed to selectively transfer knowledge and integrate the advantages of multiple VFMs. 
  Experimental results show that AIO-Stereo achieves start-of-the-art performance on multiple datasets and ranks $1^{st}$ on the Middlebury dataset and outperforms all the published work on the ETH3D benchmark.
\end{abstract}

%

\section{Introduction}
\label{sec:intro}
With the development of 3D vision tasks and their applications such as robotics and autonomous driving, stereo matching~\cite{crestereo,psmnet} becomes a fundamental vision task since it can provide depth information in the real 3D world. Stereo matching models typically predict the pixel-wise displacement (\emph{i.e.}, disparity) between a pair of rectified images and further decode the depth information with the camera calibration.

Benefiting from the success of deep learning, some works begin to explore learning-based methods~\cite{gcnet,aanet}. As a milestone, PSMNet~\cite{psmnet} utilizes 3D convolution to regularize a 4D cost volume and boost the performance. However, such learning-based methods need large computational costs. Recently, iterative optimization-based methods~\cite{crestereo,igevstereo,zhao2023high} have shown great potential on stereo matching tasks by progressively updating the disparity map. 
Selective-Stereo~\cite{wang2024selective} proposes selective recurrent unit and contextual spatial attention module to further improve the ability to predict detailed areas.


\begin{figure}[t]
  \centering
   \includegraphics[width=0.95\linewidth]{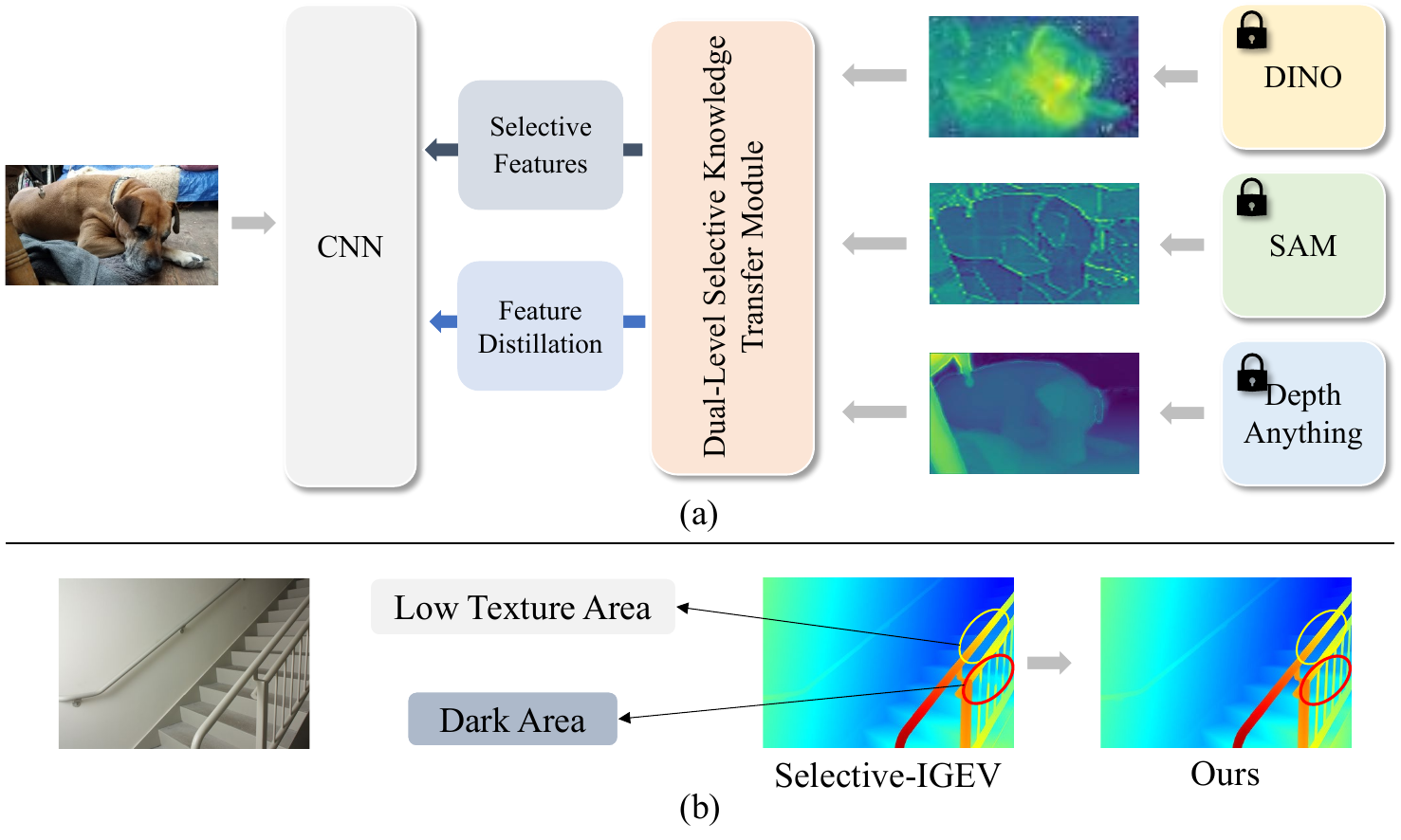}
   \caption{(a) The overview of AIO-Stereo which transfers selected knowledge from multiple VFMs to a single stereo matching model. (b) Comparisons between Selective-IGEV and our AIO-Stereo in dark and low texture areas.}
   \label{fig:fig_1}
\end{figure}

Despite the great improvement in performance, the general feature extraction capabilities of the existing models are relatively weak which can be attributed as follows: (1) Recent state-of-the-art (SOTA) works (\emph{e.g.}, Selective-Stereo~\cite{wang2024selective}) mainly focus on refining iterative update mechanisms while partially ignoring the quality of encoded features. Meanwhile, the task-oriented optimization objectives also make it difficult for the encoder to learn global and contextual information. (2) The amount of stereo matching data is relatively small and most of them are synthetic data. It is difficult for models to learn general representations from insufficient data. For example, as shown in Fig.~\ref{fig:fig_1} (b), previous methods (\emph{e.g.}, Selective-IGEV~\cite{wang2024selective}) fail to predict the depth information in dark areas with vague texture features since features in these areas are highly similar. As a result, disparity prediction based on feature matching between pixels is prone to have large mismatches.
Recently, vision foundation models (VFMs) have emerged and shown promising performance on various tasks. These VFMs are trained on large-scale datasets and can extract general representations which motivates us to consider injecting the general feature extraction capabilities of multiple VFMs into the stereo matching models.  

However, directly transferring knowledge from multiple VFMs to the single stereo matching model is not easy which is mainly caused by the following two reasons. (1) Most of the existing VFMs are based on Transformer-architecture while the stereo matching models are often based on CNN. The heterogeneity of model architectures will lead to feature mismatch when simply merging or distilling the intermediate features. (2) Different VFMs have different attention to feature representations due to the various training data, methods, and tasks. For example, as shown in~\ref{fig:fig_1} (a), DINO~\cite{caron2021emerging,oquab2023dinov2} which are pre-trained in a self-supervised manner, tend to extract global semantic information. In contrast, large segmentation models~\cite{sam,wang2023seggpt} represented by SAM~\cite{sam} pay more attention to capturing the semantic information of small objects and edges. As a result, directly using the features of multiple vision foundation models without selecting will cause feature conflicts. 

Based on the observation and analysis, we claim that the quality of encoded features is equally crucial for the stereo matching task, as they constitute the main sources of information for the iterative update modules, directly influencing every step of the iteration process. To this end, we propose an efficient knowledge transfer framework, named AIO-Stereo, that sifts and learns advantageous knowledge from multiple VFMs to obtain sufficiently effective and informative features. 
To transfer the knowledge from heterogeneous VFMs effectively and take full advantage of different VFMs,
we develop a dual-level knowledge utilization module to bridge the gap between misaligned features and transfer multi-level knowledge.
Furthermore, considering that the features derived from multiple VFMs are vastly divergent and potentially conflicting, a dual-level selective knowledge transfer module is proposed to selectively transfer knowledge and fully leverage the strengths of each VFM.
Our contribution can be summarized as follows:
\begin{compactitem}
    \item To enhance the general understanding of stereo networks, we first propose to leverage the diverse and general knowledge of multiple vision foundation models for stereo matching. 
    \item We proposed a flexible knowledge transfer framework, named AIO-stereo, which consists of dual-level knowledge utilization and a selective knowledge transfer module that can effectively and efficiently transfer the multi-level knowledge
    from multiple heterogeneous vision foundation models to a single stereo matching model.
    \item Experimental results show that the proposed AIO-Stereo ranks $1^{st}$ on the Middlebury dataset and outperforms the published methods on the ETH3D benchmark. 
\end{compactitem}

\section{Related Work}

\subsection{Stereo Matching}

As a difficult pixel-level 3D task, stereo matching has been studied for a long time and early works primarily utilize traditional matching algorithms~\cite{boykov2001fast,klaus2006segment,sun2003stereo,yang2008stereo,hirschmuller2002real,van2002hierarchical,hirschmuller2005accurate}. Since Zbontar and LeCun~\cite{zbontar2015computing} first introduced CNN to calculate the matching cost, traditional matching algorithms have gradually been replaced by learning-based methods~\cite{gcnet,aanet,mayer2016large}. 
PSMNet~\cite{chang2018pyramid} incorporates contextual information with 3D convolution and used feature concatenation to construct 4D cost volume. HITNet~\cite{tankovich2021hitnet} 
proposes a fast multi-resolution initialization step, differentiable 2D geometric propagation, and warping mechanisms that take both speed and accuracy into account. 
More recently, iterative optimization-based methods~\cite{raft-stereo,igevstereo} have shown great potential in stereo matching tasks. Inspired by~\cite{raft}, RAFT-Stereo~\cite{raft-stereo} first explores the iteration of multi-scale update blocks to generate the final disparity map from coarse to fine. IGEV-Stereo~\cite{xu2023iterative} applies additional Geometry Encoding Volume to supplement the missing non-local geometry knowledge. CREStereo~\cite{crestereo} designs an adaptive group correlation layer 
and releases a new large-scale, high-quality synthetic dataset. 
Selective-Stereo~\cite{wang2024selective} proposes a selective recurrent unit and a contextual spatial attention module that can better capture details. However, current works mainly focus on designing the iterative process and relatively ignore the feature extraction ability of encoders, which is also important in the stereo matching task. 

\subsection{Vision Foundation Models}

In recent years, thanks to the improvement of hardware performance and the construction of large-scale datasets, vision foundation models (VFMs) with extremely high performance have emerged. These VFMs can process and understand image or video information effectively and facilitate the development of other vision tasks. In image-level tasks, for example,  CLIP~\cite{radford2021learning} which is trained on image-text pairs shows a strong zero-shot classification capability. In pixel-level tasks, Depth Anything family~\cite{depth_anything_v1,depth_anything_v2} demonstrates a strong generalization ability in different depth estimation scenarios.  SAM~\cite{sam} explores a semi-supervised pipeline and achieves promising object category-agnostic segmentation capabilities. Besides, lots of studies~\cite{oquab2023dinov2,darcet2023vitneedreg,liu2021swin} introduce more general backbone networks through pre-training. As a representative, DINO family~\cite{caron2021emerging,oquab2023dinov2} explores self-supervised learning on vision transformer and boosts the performance on various downstream tasks.
Although these VFMs have shown great generalization and zero-shot ability, how to effectively utilize them to improve the performance of stereo matching still remains unexplored.

\begin{figure*}[t]
  \centering
   \includegraphics[width=0.95\linewidth]{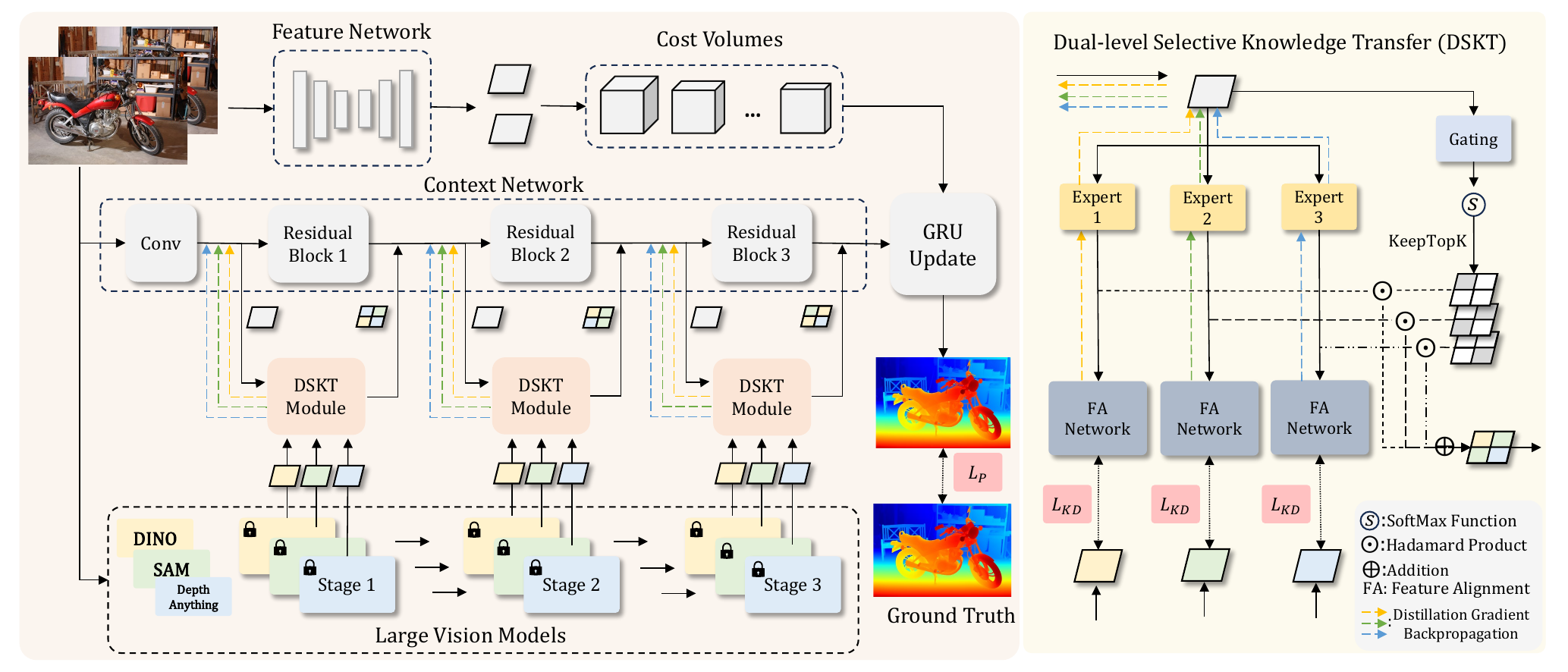}
   \caption{The Overall framework of AIO-Stereo. Left: AIO-Stereo selectively learns knowledge from SAM, DINO and Depth Anything by the proposed dual-level selective knowledge transfer module. Right: The detailed structure of our proposed dual-level selective knowledge transfer module.}
   \label{fig:fig_2}
\end{figure*}

\subsection{Knowledge Distillation}

Knowledge distillation (KD) was first proposed by Hinton~\cite{hinton2015distilling} and has been widely used in model compression~\cite{kim2018paraphrasing,bai2020few} and knowledge transfer~\cite{komodakis2017paying}. Recently, with the expansion of model zoos and the development of large models, learning knowledge from multiple teachers and heterogeneous teachers has attracted increasing attention. Following the trend, researchers began to study multi-teacher distillation~\cite{mehak2018knowledge,fukuda2017efficient} and distillation from heterogeneous architectures~\cite{shen2019customizing,touvron2021training,hao2024one}. 
To take advantage of multiple teachers, FEED~\cite{park2019feed} and Knowledge Flow~\cite{liu2019knowledge} add non-linear transformations to align the features between the student model and multiple teacher models. Besides, since models of different architecture (\emph{i.e.}, CNN, ViT, MLP) have their own distinct inductive bias, directly distilling knowledge between heterogeneous models will result in significant degradation of the performance. To handle such a problem, OFA-KD~\cite{hao2024one} transfers the mismatched feature representations into the aligned logits which contain less architecture-aware information. In this paper, we explore the knowledge transfer problem from multiple VFMs to a single stereo matching model for the first time.

\section{Method}
In this section, we first introduce the VFMs we utilized. Then, we analyze the challenges in transferring abundant knowledge from multiple heterogeneous VFMs to single stereo matching model. Finally, we detail our AIO-Stereo, which can flexibly transfer the required knowledge from multiple VFMs for the stereo matching task.

\subsection{Preliminaries: VFMs}
\label{method:preliminaries}
\textbf{DINO}~\cite{caron2021emerging,oquab2023dinov2} is built upon the Vision Transformer (ViT)~\cite{dosovitskiy2020image} architecture. 
In the training process, DINO aligns the feature representations of the same image with different augmentation methods in a self-supervised learning manner. As a result, DINO is encouraged to develop invariant and robust feature representations, especially the foreground areas and salient regions of the input image.

\textbf{SAM}~\cite{sam} is an image segmentation model that includes a ViT-based image encoder, a prompt encoder, and a lightweight mask decoder. The model is trained on a large amount of semi-supervised data in which the unlabeled data is automatically annotated by the model. Thus, features in SAM demonstrate strong zero-shot generalization to unfamiliar objects and images. Besides, as a segmentation model, SAM can extract abundant representations of diverse objects and edges and generate the segmentation mask in any location of the image according to the prompt.

\textbf{Depth Anything} family~\cite{depth_anything_v1,depth_anything_v2} is a series of VFMs for monocular depth estimation, leveraging the Transformer architecture. The model is trained on a wide variety of supervised and extensive unlabeled data, which captures depth information at multiple scales. Therefore, Depth Anything exhibits impressive generalization abilities across diverse data. It adeptly captures subtle visual cues to differentiate depth variations between objects and their surroundings, especially in dark and low-texture areas.

\subsection{Challenges of Transferring Knowledge from Multiple Heterogeneous Vision Foundation Models}
\label{method:challenges}

\subsubsection{Heterogeneity between Stereo Matching Models and various vision foundation models}

Based on the characteristics of the stereo matching task, as well as the considerations of computation cost and inference speed, most of the stereo matching models are built upon CNN structures, while vision foundation models predominantly adopt the Transformer architecture. As revealed in~\cite{hao2024one}, features from heterogeneous models reside in different latent spaces. For example, features in CNN have a strong inductive bias of locality and spatial invariance, while features in Transformer represent more global and contextual information with the self-attention module. Consequently, simply merging or distilling features between vision foundation models and stereo matching models
is unsuitable and could potentially prevent the feature learning of the stereo matching network. In the experiment process, it has a negative impact (\emph{i.e.}, -0.71 in terms of EPE) on our results and causes unstable performance.

\subsubsection{Knowledge Discrepancies and Conflicts Among Vision Foundation Models}

Despite the strong generalization and semantic understanding capability of VFMs, they still have their independent characteristics due to different training methodologies and optimization objectives. As mentioned above, DINO pays more attention to the foreground areas of objects. SAM is better at extracting features of small objects and edge areas by training on the segmentation tasks.
Depth Anything, the large monocular depth estimation model trained on various datasets, achieves superior generalization across data from different sources. Thus, features in the model excel at discerning subtle visual cues to infer the relative depth changes on the surfaces of objects.

Therefore, each VFM pays attention to different areas of the image and there exist differences and conflicts between features and knowledge from different VFMs as shown in Fig.~\ref{fig:fig_1}. As a result, indiscriminate acceptance of knowledge from multiple VFMs can lead not only to potential interferences among the disparate sources of knowledge but might also impede the model’s learning process and convergence.

\subsection{AIO-Stereo}
\label{method:aio-stereo}
In this section, we detail AIO-Stereo, a simple but effective method that can transfer knowledge from multiple VFMs flexibly and selectively. To take advantage of VFMs, a dual-level knowledge utilization is designed to effectively transfer knowledge between heterogeneous models, and a selective knowledge transfer module is proposed to integrate knowledge from multiple VFMs into a single stereo matching model which will be introduced later.

\subsubsection{Overview of AIO-Stereo}
As shown in Fig.~\ref{fig:fig_2}, the feature extractor can be divided into a feature network to calculate cost volumes and a context network to generate context features for GRU refinement. Specifically, the context network contains three residual blocks with each block consisting of a series of residual and downsampling layers. Our dual-level selective knowledge transfer (DLSKT) module is adapted to transfer the rich knowledge of the vision foundation models to the context network. In detail, given the left images or the right images {\small $I^{l(r)} \in \mathbb{R}^{3 \times H \times W }$} as inputs, local features $c^{l(r)}$ are extracted by the feature network and the pixel-wise correlation volume can be calculated by:
\begin{equation}
Corr(x,y,z) =  \langle c^l(x,y), c^r(x - z,y) \rangle,
\end{equation}
where {\small $\langle \cdot,\cdot \rangle$} is the inner product. Meanwhile, context features can be obtained by the context network taking the left images {\small $I^{l}$} as the input. In this process, a single layer convolution with a kernel size of 7 is first applied to the image to get the original features {\small $f_{0} \in \mathbb{R}^{{C}_{0} \times H \times W}$}, where {\small ${C}_{0}$} is the number of channels. Then, the original features are fed into the residual blocks, with knowledge gradually learned from the VFMs by the DLSKT module. Specifically, a series of intermediate features {\small ${f_{i} \in \mathbb{R}^{{C}_{i} \times \frac{H}{2^{i-1}} \times \frac{W}{2^{i-1}} } }(i=1,2,3)$} can be obtained, where {\small ${C}_{i}$} is the feature channels. For the {\small $i^{th}$} block, {\small $f_{i-1}$} is taken as the input, and {\small $f_{i}$} is the output feature.  
Following RAFT-Stereo~\cite{raft-stereo}, correlation volumes at four resolutions are obtained by average pooling. Since the context features of the left image and the right image are semantically similar, the correlation volumes and the context features of the left image are injected into every step of the GRU updating operator to refine the disparity map step by step. With a series of intermediate predictions {\small $\{\mathbf{p}_{i}\}_{i=1}^{N}$}, the prediction loss can be calculated following ~\cite{raft-stereo}:
\begin{equation}
\label{pred_loss}
L_P = \sum_{i=1}^{N} \gamma^{N-i}_{P} ||\mathbf{p}_i-\mathbf{p}_{GT}||_1,
\end{equation}
where {\small $\gamma_{P}$} is the decay factor and {\small $\mathbf{p}_{GT}$} is the ground truth.

\subsubsection{Dual-Level Knowledge Utilization}
\label{dual-level}
To better integrate knowledge from the heterogeneous VFM into our stereo matching model, we propose dual-level knowledge utilization which utilizes the knowledge in both distillation and fusion levels. For simplicity and better understanding, we take a single large model, SAM, as an example to show our knowledge utilization method. As mentioned above, since VFMs and our conv-based backbone are heterogeneous, direct feature distillation or feature fusion is not appropriate since the features are in different latent spaces and will result in misalignment. To bridge the architectural gap between SAM and the CNN backbone, we integrate an expert network and a feature alignment network to align the features between heterogeneous models and transfer general representations from the VFM to the stereo matching model. Specifically, a lightweight expert network $E_{i}^{s}$ is designed and takes the intermediate features $f^{}_i$ as the input to get the expert features $e^{s}_{i}$ which can be calculated as follows: 
\begin{equation}
e^{s}_{i} = E_{i}^{s}(f^{}_i  \;|\; \varphi_{i}^{s}),
\end{equation}
where $\varphi_{i}^{s}$ denotes the parameter of the expert network. Then, the expert features $e^{s}_{i}$ subsequently pass through a heavier feature alignment network $A_{s}^{i}$ which reconciles the feature space of the stereo network with that of SAM. A distillation loss is finally designed to make the stereo matching model learn from the VFM in the aligned feature space which can be written as follows:
\begin{equation}
\label{KD_loss}
L_{KD,i}^{s} = \mathbf{MSE}(A_{i}^{s}(e^{s}_{i} \;|\; \theta_{i}^{s}) , \mathcal{F}^s(s^{}_i)),
\end{equation}
where $s^{}_i$ represents the features extracted from the $i^{th}$ stage of SAM, $\mathcal{F}^s$ denotes the interpolate function to align the resolutions, and $\theta_{i}^{s}$ is the parameter of feature alignment network. During the distillation process, the knowledge derived from SAM is propagated to the expert and the backbone network through the feature alignment network. A heavier feature alignment network can reduce the misalignment among features, but may also lead to greater knowledge attrition. Thus, we initialize the feature alignment network with a higher learning rate to rapidly learn the mapping relationships within the feature space and a larger learning rate decay factor to prevent excessive retention of knowledge within the network. Through the distillation process, knowledge is conveyed to the preceding blocks by backpropagation, while the forward propagation of the current block may attenuate the knowledge to some extent. To enhance the knowledge, we add the output of the expert network back into the output of the of the current residual block:
\begin{equation}
f^{}_{i+1} = B_{i}(f^{}_i \;|\; \zeta_{i}) + e_{i}^{s},
\end{equation}
where $B_{i}$ is the residual block with $\zeta_{i}$ to be its parameter. This dual-level design ensures a more effective knowledge transfer from the heterogeneous vision foundation model.

\subsubsection{Selective Knowledge Transfer}
\label{skt}
To facilitate effective knowledge transfer from multiple distinct VFMs while preventing knowledge interference between different models, we design a dual-level selective knowledge transfer module. Inspired by the mechanism of the mixture of experts (MoE)~\cite{6797059}, we employ a selective knowledge transfer mechanism with a trainable gating network to select the experts for the features of each pixel. Specifically, for the $i^{th}$ block, we extract the features from the $i^{th}$ stage of DINO, SAM, and Depth Anything, represented as $d^{}_{i}$, $s^{}_{i}$, and $a^{}_{i}$ respectively. As shown in Fig.~\ref{fig:fig_2}, each VFM is associated with an expert network that learns the knowledge from the corresponding VFM through distillation. Taking $f^{}_i$ as inputs, $e^{d}_{i}$, $e^{s}_{i}$, $e^{a}_{i}$ can be obtained by the expert networks $E^{d}_{i}$, $E^{s}_{i}$, and $E^{a}_{i}$ respectively. The distillation loss for multiple VFMs can be calculated as follows:

\begin{equation}
\begin{aligned}
    L_{KD,i} &= \sum_{x \in \{d,s,a\}}L_{KD,i}^{x} 
    \end{aligned}
\end{equation}

\begin{equation}
g^{}_i = \text{KeepTopK}(\text{Softmax}(G_{i}(f^{}_i) \;|\; \psi_{i}),k,dim=0),
\end{equation}
where $k$ is the number of experts to be retained at each pixel, and $\text{KeepTopK}(\cdot,k,\cdot)$ is a function that keeps the top-$k$ values with the highest weight at the specific dimension. Subsequently, we use selection weights to determine the importance of the features from three experts, selectively fusing and discarding certain features. Finally, the selected features are then added to the output of the current residual block as the input of the next block.
\begin{equation}
f^{}_{i+1} = B_{i}(f^{}_i \;|\; \zeta_{i}) + \sum_{x \in \{d,s,a\}} e^{x}_{i} \odot g^{}_i(x)
\end{equation}
By selecting the features from experts, we indirectly choose the knowledge from multiple VFMs at different regions of the image to transfer the most beneficial and effective knowledge from each large model.

\subsubsection{Loss Function}
The overall loss function consists of the prediction loss $L_P$ and the distillation loss $L_{KD}$. The L1 Loss is calculated between the predicted disparity maps and the ground truth. And the MSE Loss is applied for knowledge distillation. The final loss is defined as:
\begin{equation}L_{AIO} = L_P+L_{KD} = L_P + \sum_{j=1}^{3}\gamma^{4-j}_{KD} L_{KD,j},
\end{equation}
where $\gamma_{KD}$ is the decay factor and $L_P$, $L_{KD,j}$ are defined in Eq.~(\ref{pred_loss}) and Eq.~(\ref{KD_loss}), respectively.

\section{Experiments}
\label{sec:experiments}

\subsection{Datasets}
Following Selective-Stereo~\cite{wang2024selective}, we verify the effectiveness of AIO-Stereo on four widely used datasets including Scene Flow~\cite{mayer2016large}, Middlebury 2014~\cite{middlebury}, KITTI-2015~\cite{kitti2015} and ETH3D~\cite{eth3d}. Scene Flow~\cite{mayer2016large} contains more than 39000 synthetic stereo frames which are divided into training and testing set. 
Middlebury 2014~\cite{middlebury} provides a training set with images of 23 indoor scenes and a testing set with images of 10 indoor scenes, and both sets have three resolutions to use. KITTI-2015~\cite{kitti2015} contains 200 training pairs and 200 testing pairs with sparse disparity maps which were collected in real-world driving scenes. ETH3D~\cite{eth3d} provides gray-scale image pairs covering both indoor and outdoor scenes.

\begin{table*}[t]
  \centering
    \begin{tabular}{l|cccccc|cc}
    \toprule
    Method & Distillation & Forward & Selection & DINOv2  & SAM & Depth Anything v2&\makecell{ EPE \\ (px) } &\makecell{ \textgreater2px \\ (\%) }\\
    \midrule
    \midrule
    Baseline & & & & & & &0.74 &4.68 \\
    \midrule
    w/o Selection &\checkmark &\checkmark & &\checkmark &\checkmark &\checkmark &0.68 &3.57 \\
    w/o Distillation & &\checkmark &\checkmark &\checkmark &\checkmark &\checkmark &0.72 &3.87 \\
    w/o Forward Fusion &\checkmark & & &\checkmark &\checkmark &\checkmark &0.67 &3.52 \\
    Only DINO &\checkmark &\checkmark &\checkmark &\checkmark & & &0.66 &3.64 \\
    DINO+SAM &\checkmark &\checkmark &\checkmark &\checkmark &\checkmark & &0.68 &3.61 \\
    \midrule
    full* &\checkmark &\checkmark &\checkmark &\checkmark &\checkmark &\checkmark & \textbf{0.66} & \textbf{3.48} \\
    \bottomrule
  \end{tabular}
  \caption{
  Ablation study of proposed networks on the MiddEval v3 training set in full resolution and all metrics are on all pixels. The baseline is the Selevtive-IGEV~\cite{wang2024selective}.* means the final version of our method. }
  \label{tab:headings}
\end{table*}

\begin{table*}[t]
    \centering
    \setlength{\tabcolsep}{1.6mm}
    \begin{tabular}{l|c c c c| c c c c|ccc}
     \toprule
     \multirow{2}{*}{Method}   &\multicolumn{4}{c|}{Middlebury} & \multicolumn{4}{c}{ETH3D}& \multicolumn{3}{c}{ KITTI 2015}\\
      &  bad1.0&bad2.0&avgerr&A90&bad0.5&bad1.0&bad2.0&avgerr& D1-bg & D1-fg & D1-all  \\
     \midrule
    PSMNet~\shortcite{psmnet}   &63.9 &42.1 &6.68 &17.0&-&-&-&-& 1.86 & 4.62 & 2.32 \\
    HITNet~\shortcite{hitnet} &13.3 &6.46 &1.71 &2.32&7.83&2.79&0.80&0.20& 1.54  & 2.72 & 1.74\\
    RAFT-Stereo~\shortcite{raft-stereo} &9.37 &4.74 &1.27 &1.10&7.04&2.44&0.44&0.18&1.75&2.89&1.96 \\
    LEAStereo~\shortcite{leastereo}   &20.8 &7.15 &1.43 & 1.68&-&-&-&-& 1.74 & 3.20 & 1.98 \\
    CREStereo~\shortcite{crestereo} &8.25 &3.71 &1.15&0.92&3.58&\underline{0.98}&\underline{0.22}&\underline{0.13}&1.45& 2.86 &1.69 \\
    GMStereo~\shortcite{xu2023unifying} &23.6 &7.14 &1.31 &1.64&5.94&1.83&0.25&0.19&1.49 &3.14 &1.77\\
    IGEV-Stereo~\shortcite{igevstereo} &9.41 &4.83 &2.89 &4.87&3.52&1.12&\textbf{0.21}&0.14&1.38 &2.67 &
1.59 \\
    DLNR~\shortcite{zhao2023high}&6.82&3.20&1.06&0.85&-&-&-&-&1.60 &\underline{2.59} &1.76 \\
    Selective-IGEV~\shortcite{wang2024selective}&\underline{6.53}&\underline{2.51}&\underline{0.91}&\underline{0.79}&\underline{3.06}&1.23&\underline{0.22}&\textbf{0.12}&\textbf{1.33} &2.61 &\underline{1.55} \\
    AIO-Stereo (Ours) &\textbf{6.08} &\textbf{2.36} &\textbf{0.85} &\textbf{0.76}&\textbf{2.91}&\textbf{0.94}&\textbf{0.21}&\underline{0.13}&\underline{1.35} &\textbf{2.46} &\textbf{1.54}\\
    \bottomrule
    \end{tabular}
    \caption{Quantitative evaluation on Middlebury\cite{middlebury}, ETH3D~\cite{eth3d}, and KITTI 2015~\cite{kitti2015}. \textbf{Bold}: Best. \underline{Underline}: Second best.}
    \label{tab:kitti}
\end{table*}

\subsection{Implementation Details}
We implement our AIO-Stereo with Pytorch framework and perform our experiments using NVIDIA A100 GPUs while using the AdamW optimizer. For pre-training, we trained our model on the augmented Scene Flow training set (\emph{i.e.}, both cleanpass and finalpass) for 200k steps with a batch size of 8, and we use a random crop size of 320 $\times$ 720. We use a one-cycle learning rate schedule with warm up strategy and the learning rate gradually increases to 0.0002 in the first 1\% of steps and gradually decreases thereafter. And for finetune, the learning rate linearly decays from 0.0003 to 0.

\subsection{Ablation Study}

\subsubsection{Ablation for Each VFM}
AIO-Stereo leverages the knowledge from three VFMs (\emph{i.e.}, DINOv2, SAM, and Depth Anything v2) to enhance feature representation and improve overall accuracy. To verify our method can effectively integrate the advantage of multiple VFMs,
we conduct experiments on using different numbers of VFMs (\emph{i.e.}, only DINO, DINO and SAM, DINO, SAM and Depth Anything). As shown in the upper part of Tab.~\ref{tab:headings}, performance improves when using VFMs which verifies that AIO-Stereo can effectively learn from VFMs. Besides, as the number of used VFMs increases, the performance is consistently improved. This is because each VFM has its unique advantages and knowledge from different VFMs can be integrated effectively by our AIO-Stereo. Moreover, the results highlight the inherent flexibility of our AIO-Stereo, which is not dependent on a single foundation model but is designed to effectively orchestrate multiple models, leveraging their strengths to serve our stereo matching task. It suggests that our AIO-Stereo can flexibly take advantage of various VFMs.

\subsubsection{Effectiveness of Dual-level Knowledge Utilization}
To effectively transfer the knowledge between heterogeneous models, we use the knowledge in both distillation and fusion levels. In this section, we evaluate the effectiveness of our dual-level approach to knowledge utilization from VFMs. In detail, we first exclude the aligned knowledge distillation to eliminate the effect of extra parameters brought by the expert networks. It can be observed from Tab.~\ref{tab:headings} that the performance decreases largely (\emph{i.e.} 3.48 to 3.87 on the 2 pixels error index) which is because the model will be unable to learn from the knowledge of VFMs without the distillation process. 
Besides, there is also a performance drop without the forward fusion. This is because the model is more prone to forgetting the knowledge it has acquired, leading to a diminished effect in knowledge transfer. By utilizing knowledge at both levels, our approach achieves a more comprehensive and rich transfer of visual knowledge.

\subsubsection{Exploration of DLSKT}
In this section, we explore the expert selection mechanism of our DLSKT module. In particular, we apply a non-selective knowledge transfer which accepts all the knowledge from VFMs indiscriminately to compare with our selective knowledge transfer. The results in Tab.~\ref{tab:headings} (\emph{i.e.}, w/o selection) indicate that the non-selective knowledge transfer underperforms our selective methods on both EPE and 2 pixels error index, attributed to the incompatible and sometimes contradictory knowledge among different VFMs. 

\subsection{Comparisons with State-of-the-art}
To evaluate the effectiveness of our method, we compare AIO-Stereo with the current SOTA methods on the Middlebury, ETH3D, and KITTI 2015 datasets as shown in Tab.~\ref{tab:kitti}. Note that AIO-Stereo ranks $1^{st}$ on the Middlebury leaderboard and achieves SOTA on multiple datasets.

\subsubsection{Middlebury.}
For the Middlebury dataset, following~\cite{wang2024selective}, we first finetune our pre-trained model on the mixed Tartan Air~\cite{wang2020tartanair}, CREStereo Dataset~\cite{crestereo}, Scene Flow, Falling things~\cite{tremblay2018falling}, InStereo2k~\cite{bao2020instereo2k}, CARLA HR-VS~\cite{yang2019hierarchical}, and Middlebury datasets 200k steps using a crop size of 384 $\times$ 512 with a batch size of 8. Then we finstune it on the mixed CREStereo Dataset, Falling Things, InStereo2k, CARLA HR-VS, and Middlebury datasets using a crop size of 384 $\times$ 768 with a batch size of 8 for another 100k steps. As shown in Tab.~\ref{tab:kitti}, our method achieves SOTA performance on the Middlebury test set. Specifically, our method surpasses Selective-IGEV~\cite{wang2024selective} and DLNR~\cite{zhao2023high} by 5.98\% and 26.25\% on the bad 2 pixels error respectively without extra design to the refinement process, demonstrating the effectiveness of our designs.

\subsubsection{ETH3D.}
For the ETH3D dataset, following~\cite{wang2024selective}, we finetune the pre-trained model on the mixed Tartan Air, CREStereo Dataset, Scene Flow, Sintel Stereo~\cite{butler2012naturalistic}, InStereo2k, and ETH3D datasets for 300k steps. Then we fintune it on the mixed CREStereo Dataset, InStereo2k, and ETH3D datasets for another 90k steps. Our method achieves the best performance among all published methods for most metrics, and outperforms Selective-IGEV~\cite{wang2024selective} by 23.58\% on bad 1.0 metric. quantitative results are shown in Tab~\ref{tab:kitti}. 

\subsubsection{KITTI-2015.}
For the KITTI-2015 dataset, following ~\cite{wang2024selective}, we finetune the pretrained model on the mixed dataset of KITTI-2012~\cite{kitti2012} and KITTI-2015 with a batch size of 8 for 50k steps. As shown in Tab.~\ref{tab:kitti}, our method achieves comparable results and surpasses Selective-IGEV by 5.75\% on D1-fg metric.

\begin{table}[t]
\centering
    \centering
    \centering
    \begin{tabular}{l|cc|cc}
    \toprule
    \multirow{2}{*}{Method} & \multicolumn{2}{c|}{F} &\multicolumn{2}{c}{H} \\
    &EPE&D1&EPE&D1\\
    \midrule
    PSMNet~\shortcite{psmnet} &40.51&57.93&9.79&32.19\\
    RAFT-Stereo~\shortcite{raft-stereo} &\textbf{3.84}&15.64&1.44&11.21\\
    IGEV-Stereo~\shortcite{igevstereo}&5.87&\underline{11.85}&1.36&\underline{7.21} \\
    GMStereo~\shortcite{xu2023unifying}&\underline{4.10}&29.15&1.92&15.69\\
    EAI-Stereo~\shortcite{zhao2022eai}&6.16&18.25&2.15&11.74\\
    DLNR~\shortcite{zhao2023high} &6.57&14.46&1.45&9.46\\
    Selective-IGEV~\shortcite{wang2024selective} &5.28&12.07&\underline{1.35}&7.31\\
    AIO-Stereo (Ours) &4.16&\textbf{11.67}&\textbf{0.89}&\textbf{6.48}\\
    \bottomrule
  \end{tabular}
  \caption{Zero-shot evaluation on Middlebury. \textbf{Bold}: Best. \underline{Underline}: Second best.}
  \label{tab:eth}
\end{table}

\subsection{Zero-Shot Generalization}
To evaluate the generalization capabilities of our proposed method, we pre-train our model on the synthetic Scene Flow dataset and directly test it on the Middlebury dataset, an unseen real-world dataset with challenging indoor scenes. As shown in Tab.~\ref{tab:eth}, AIO-Stereo achieves state-of-the-art performance at most of the metrics. Attributed to the knowledge transferred from the vision foundation models, our method performs well on unseen environments.


\subsection{Visualization}

\begin{figure}[t]
    \includegraphics[width=0.97\linewidth]{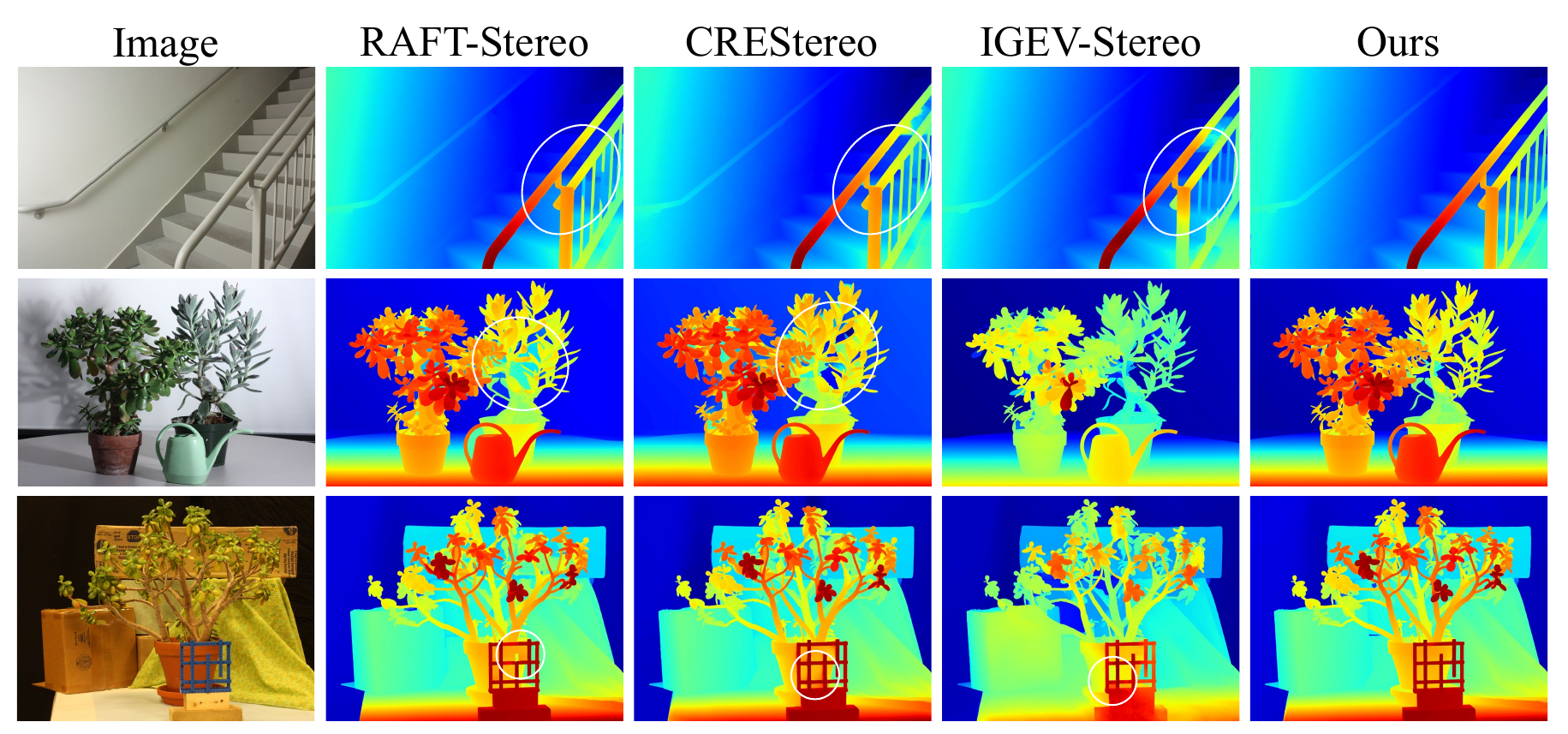}
   \caption{Visual comparison on the Middlebury dataset.}
   \label{fig:fig_4}
\end{figure}

\subsubsection{Visual Comparisons on Middlebury} 
Further, we compare the visualization results with other works (\emph{i.e.}, RAFT-Stereo~\cite{raft-stereo}, CREStereo~\cite{crestereo}, and IGEV-Stereo~\cite{igevstereo}). It can be seen from Fig.~\ref{fig:fig_4} that AIO-Stereo can achieve better visualization quality, especially in texture and dark areas. This is because our method can take advantage of multiple VFMs and learns general representations from them.  

\begin{figure}[t]
  \centering
   \includegraphics[width=0.94\linewidth]{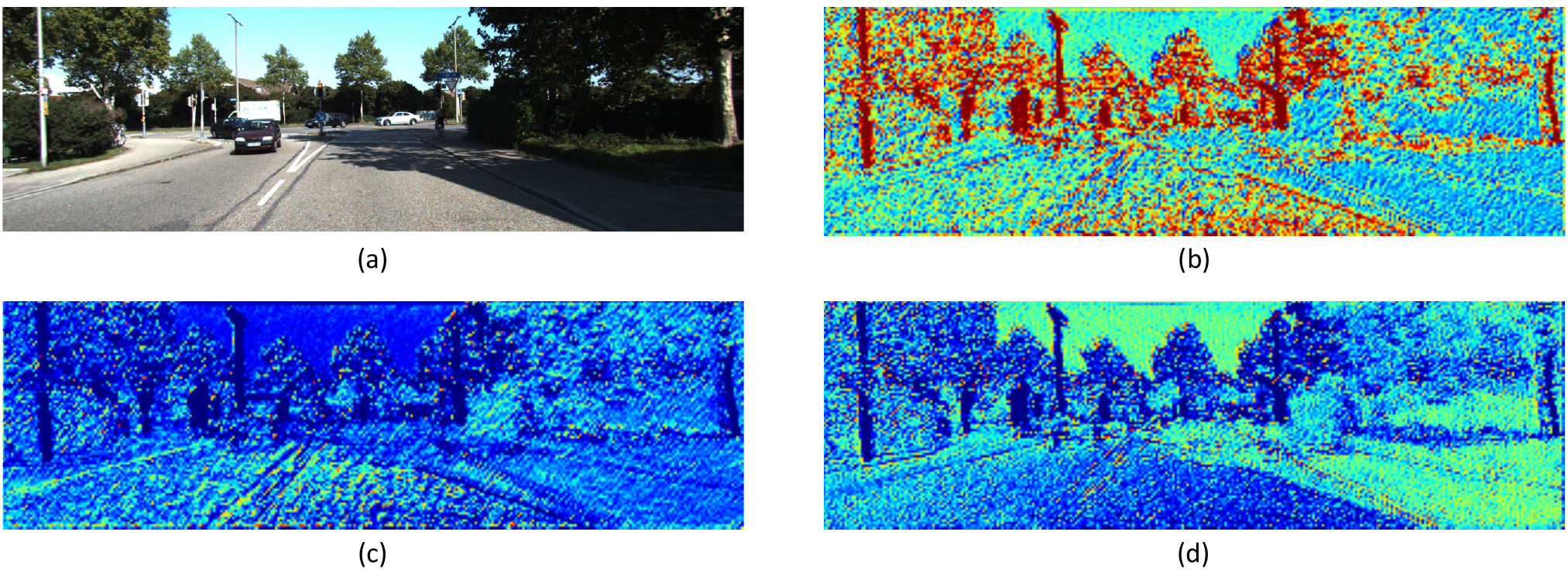}
   \caption{Visualization of the selection weights for each VFM. (a) Reference image. (b-d) Selection weights of DINO, SAM, and Depth Anything respectively.}
   \label{fig:fig_3}
\end{figure}

\subsubsection{Visualization of Selection Weights for Each Expert} Our method indirectly selects different VFMs by selecting different experts. We visualize the selection weight for each VFM in Fig.~\ref{fig:fig_3} to show our strategy of selecting from different regions based on the independent strengths of various models. The visualization indicates a clear preference for foreground regions when integrating features from DINO, attributable to its enhanced focus and robustness within these regions. For SAM, features are mainly selected on the edges, aligning with SAM’s capability for identifying differences between objects. As for Depth Anything, our model learns features of dark and low-texture areas, where Depth Anything performs well. The visualization results further verify that AIO-Stereo can combine the advantage of different VFMs.


\section{Conclusion}
\label{conclusion}
In this paper, for the first time, we explore leveraging the knowledge of VFMs to improve the performance of stereo matching. Specifically, we propose AIO-Stereo which can transfer knowledge from multiple VFMs into a single stereo matching model. Our AIO-Stereo combines the advantages of multiple VFMs by selective knowledge transfer module and effectively adapts the knowledge from heterogeneous VFMs to our stereo matching model by dual-level knowledge utilization module. Experimental results show that our AIO-Stereo achieves SOTA performance on multiple datasets and rank $1^{st}$ on the Middlebury dataset. 

\section{Acknowledgments}
This work is supported by Shanghai Natural Science Foundation (No. 23ZR1402900), National Key Research and Development Program of China (No. 2022ZD0160101), Shanghai Municipal Science and Technology Major Project (No.2021SHZDZX0103).
The computations in this research were performed using the CFFF platform of Fudan University.


\bibliography{aaai25}

\end{document}